\documentclass[12pt]{spieman}  
\usepackage{amsmath,amsfonts,amssymb}
\usepackage{graphicx}
\usepackage{setspace}
\usepackage{tocloft}
\usepackage{amsmath}
\usepackage{epsfig}
\usepackage{amssymb}
\usepackage{multirow}
\usepackage{algorithm}
\usepackage{algpseudocode}
\usepackage{balance}
\usepackage{xcolor}
\usepackage{lineno}
\usepackage{rotating}
\usepackage{multirow}
\usepackage{lineno}

\title{Detection of River Sandbank 
for Sand Mining 
with the Presence of Other High Mineral Content Regions 
Using Multi-spectral Images}

\author[*]{Jit Mukherjee}
\affil[*]{Dept. of Computer Science and Engineering, Birla Institute of Technology, Ranchi, Jharkhand, India}

\cftpagenumbersoff{figure}
\cftpagenumbersoff{table} 
\begin{document} 
\maketitle

\begin{abstract}
Sand mining is a booming industry. The river sandbank is one of the primary sources of sand mining.
Detection of potential river sandbank regions for sand mining directly impacts the economy, society, and environment. 
In the past, semi-supervised and supervised techniques have been used to detect mining regions including sand mining. 
A few techniques employ multi-modal analysis combining different modalities such as multi-spectral imaging, synthetic aperture radar (\emph{SAR}) imaging, aerial images, and point cloud data. 
However, the distinguishing spectral characteristics of river sandbank regions are yet to be fully explored.
This paper provides a novel method to detect river sandbank regions for sand mining using multi-spectral images without any labeled data over the seasons. 
Association with a river stream and the abundance of minerals are the most prominent features of such a region.
The proposed work uses these distinguishing features to determine the spectral signature of a river sandbank region, which is robust to other high mineral abundance regions.
It follows a two-step approach, where first, potential high mineral regions are detected and next, they are segregated using the presence of a river stream.
The proposed technique provides average accuracy, precision, and recall of $90.75\%$, $85.47\%$, and $73.5\%$, respectively over the seasons from Landsat 8 images without using any labeled dataset.
\end{abstract}


{\noindent \footnotesize\textbf{*}Jit Mukherjee,  \linkable{jit.mukherjee@bitmesra.ac.in} }


\section{Introduction}
\label{sec:intro}
River sand is a widely used construction material with high economic and ecological value.
Rapid urbanization in several developing countries such as India, has increased its requirement leading to uncontrolled and illicit mining impacting the environment. 
Excavation of sand from river sandbank has a few advantages as it can avoid flood inundation to a certain degree.
However, the abundance of sand mining and illicit, unmonitored sand quarries create disastrous circumstances due to their effects on river morphology, riverbed stability, change in the water table, erosion, surrounding eco-system, and habitation~\cite{Stalin2020,WANG2012340,Mngeni_efectSandMine_2016}.
As, in the natural process, sand formation takes over a few hundred years, monitoring and detection of potential sand mining regions have several high impacts on the environment.
Earlier, geodetic surveys, which are tedious, are used for monitoring and measurement of excavated sands.
Satellite imaging is found to be instrumental for land class detection and monitoring remotely.
In recent advances, different modalities of remote sensing are used in various aspects of mining including sand mining~\cite{Stalin2020,indriasari2018identification,jit_mtap_overburden}. 
\subsection{Related Works}
Remote sensing applications in sand mining have multifold research challenges such as detection of illicit mining, monitoring, volumetric control, etc.~\cite{Stalin2020,Mngeni_efectSandMine_2016}.
In the literature, 
multimodal analysis is employed for such applications mostly using labelled data.
In~\cite{indriasari2018identification}, a multi-modal analysis composed of multi-spectral and interferometric synthetic aperture radar (\emph{InSAR}) images is used to quantify ex-sand mining areas and related land deformation. 
The study is divided into two parts viz. detection of sand mining regions 
by the top of atmosphere (\emph{TOA}) reflectance 
using Landsat 5 and 8 images and thereafter, computation of the digital elevation model for quantification of land deformation. 
Delineation of the Tasseled cap transformation's scatter plot is studied 
for the identification of ex-sand mine areas
~\cite{indriasari2018identification}. 
Integration of differential global positioning systems and unmanned aerial vehicles (\emph{UAV}) has been considered to monitor and quantify sand quarry volumes~\cite{Stalin2020}. 
Mining regions are separated through coordinate points and images using a supervised deep learning technique in~\cite{balaniuk2020mining}.
These techniques focus more on detecting mining regions using a marked dataset rather than defining the distinguishing characteristics of a sand mining region.
\par
There is a significant research gap to identify river trails and sandbank regions using multi-spectral images exclusively.
Moreover, most of these techniques does not focus on the surrounding regions of a sand mine.
Detection of river trails can be found instrumental to detect such regions.
Digital elevation model (\emph{DEM}) data has been studied to detect drainage and different topographic patterns~\cite{wu2019high, hosseinzadeh2011drainage}.
An accurate \emph{DEM} generation requires prerequisites of a short temporal baseline, suitable perpendicular baseline, and atmospheric conditions.
Processed \emph{DEM} images such as advanced spaceborne thermal emission and reflection radiometer (\emph{ASTER}), and shuttle radar topography mission (\emph{SRTM}) may not be acquired in regular intervals in most of the cases.
Multi-spectral images, e.g., Landsat $8$, are obtained in a regular interval which is crucial for monitoring such a dynamic land class.
Spectral indexes are widely used in remote sensing such as normalized difference vegetation index~\emph{(NDVI), normalized difference water index~\emph{(NDWI)}~\cite{gao1996ndwi}, bare soil index~\emph{(BI)}}, etc.
In a few works, characteristics of spectral indexes such as bare soil index (\emph{BI}) have been exploited to identify high mineral content regions such as river sandbanks~\cite{jit_mtap_overburden}.
Hydrothermally altered mineral assemblages near mine water bodies have been mapped using clay mineral and iron oxide ratio~\cite{jitm2018igarss_extended, jitm2018igarss}.
However, the performances of these indexes in sand mining regions are yet to be fully explored.
\subsection{Objectives}
Semi-supervised or supervised detection of mining land classes has the prerequisite of a labeled dataset. 
Moreover, a few of them employ multi-modal analysis and \emph{DEM} data, which can be difficult to avail in certain scenarios.
Distinguishing spectral characteristics for the detection of river sandbank regions through geophysical and spectral indexes are yet to be explored.
In the literature, the detection of river sandbank regions with the presence of other regions with high mineral abundance has not been thoroughly discussed.
Thus, the objective of this paper is to detect river sandbank regions without any labeled dataset by investigating the spectral characteristics of river sandbank and their surroundings using multi-spectral images exclusively.
Furthermore, it aims to provide a technique robust to seasonal variation and other land classes with high mineral abundance.
The proposed technique can be found beneficial in different aspects of the environmental monitoring and sand mining industry.
\section{Background Techniques}
Background techniques, which are used in the proposed method are described below.
\subsection{Modified Normalized Difference Water Index}
Many techniques and indexes are proposed to detect water bodies, such as Normalized Difference Water Index (\emph{NDWI}), modified normalized difference water index (\emph{MNDWI}), Automated Water Extraction Index (\emph{AWEI})~\cite{feyisa2014awei}, etc. 
\emph{NDWI} is defined as $\frac{\lambda_{NIR}-\lambda_{SWIR-I}}{\lambda_{NIR}+\lambda_{SWIR-I}}$. 
Another variant of \emph{NDWI} is also proposed as $\frac{\lambda_{Green}-\lambda_{NIR}}{\lambda_{Green}+\lambda_{NIR}}$~\cite{mcfeeters1996ndwi, gao1996ndwi}.   
In this work, modified normalized difference water index (\emph{MNDWI}) has been used because it enhances the open water features and remove build-up noises~\cite{xu2006mndwi}.
\emph{MNDWI} is proposed as a spectral index of green and short wave infra-red band i.e. $\frac{\lambda_{Green}-\lambda_{SWIR-I}}{\lambda_{Green}+\lambda_{SWIR-I}}$.
Higher values of \emph{MNDWI} enhance water body regions.
Here, $\lambda_{Green}$, $\lambda_{NIR}$, and $\lambda_{SWIR-I}$ are defined as the spectral reflectance values of green, near infra-red, and short wave infra-red one bands, respectively.
\subsection{Coal Mine Index (\emph{CMI})}
Coal mine index (\emph{CMI}) is defined as a spectral ratio of short wave infra-red band one and short wave infra-red band two, i.e $\frac{\lambda_{SWIR-I}-\lambda_{SWIR-II}}{\lambda_{SWIR-I}+\lambda_{SWIR-II}}$.
Here, $\lambda_{SWIR-I}$, and $\lambda_{SWIR-II}$ are defined as the reflectance of short wave infra-red bands one and two, respectively.
Lower values of \emph{CMI} have been used to detect surface coal mining regions in~\cite{jit_jstar_coal_mine}.
It has been observed that regions of high mineral contents can be detected using a threshold over \emph{CMI}~\cite{jit_mtap_overburden}. 
\subsection{Connected Component Labeling}
Connected component labeling or analysis is an image processing technique to detect connected regions mostly in binary images.
It can be also applied for higher-dimensional data.
In connected component analysis, the whole image is scanned for each unmarked pixel. 
In this scanning process, it gathers all similar pixels based on pixels connectivity.
\subsection{Morphological Dilation \& Image Thinning}
\label{sec:image_thinning}
Morphological dilation and erosion are the two basic morphological operations in image processing.
In dilation, a structural element is used to probe and expand the shapes in an image. 
It increases the size of an object and fills holes. 
Areas, which are separated by space smaller than the structural element, get connected using this technique. 
Morphological dilation is represented as $A\oplus B$, where A is an image object and B is a structuring element.
Image thinning, a morphological operation, is applied on binary images to 
obtain the skeleton structures of different shapes by removing black foreground pixels layer by layer. 
Image thinning is used here for the skeletonization of detected water bodies.
Skeletonization is the process of creating a thinned version of a shape, which is equidistant to the boundaries.  
As image thinning uses a hit-or-miss transform, it preserves the shape topology.
The algorithm proposed in~\cite{zhang1984fast} has been used here for thinning the outcome of the dilation.
\section{Methodology}
\label{sec:method}
\begin{figure}[ht!]
	\centering
	\includegraphics[width=\columnwidth]{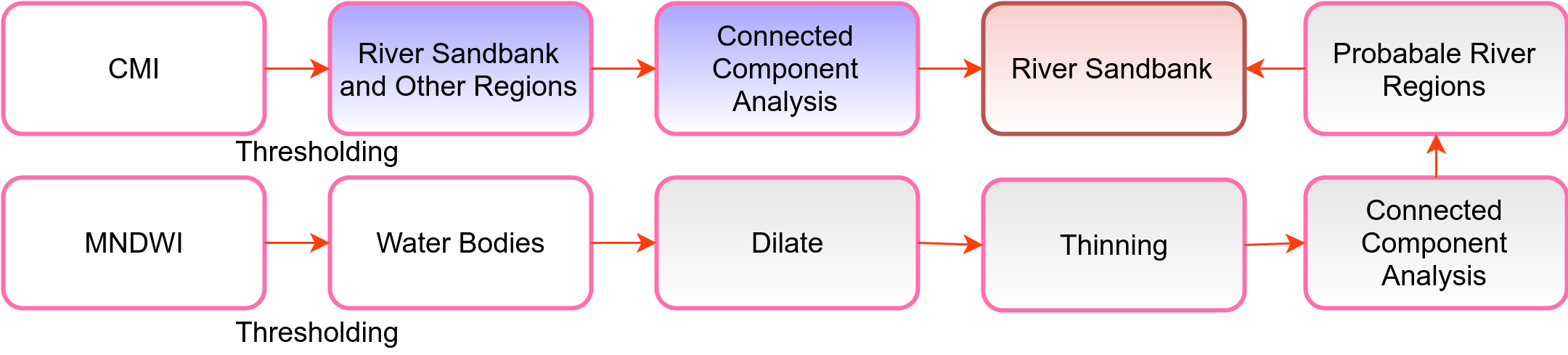}
	\caption{Flow Diagram of the Proposed Method}
	\label{fig:flow}
\end{figure}
Sand mining on river sandbanks occurs within the proximity of a river. 
This characteristic is thematically used in the proposed work to detect probable river sandbank regions with high mineral abundance.
The proposed technique consists of two segments as shown in Fig.~\ref{fig:flow}. 
First, the probable river regions are detected. 
Next, regions with high mineral quantity including river sandbank regions are marked. 
Further, the regions with high mineral abundance, which are close to a river, are separated from other high mineral abundance regions~\footnote{A version of code and a few associated data can be accessed at https://github.com/MitJukherjee/River\_Sandbank}.
\subsection{Detection of Probable River Regions}
In this work, the modified normalized difference vegetation index (\emph{MNDWI}, $\frac{\lambda_{Green}-\lambda_{SWIR-I}}{\lambda_{Green}+\lambda_{SWIR-I}}$)
is used to detect water bodies as it enhances the open water features and removes build-up noises~\cite{xu2006mndwi}.
As a river region with a prominent river sandbank can get narrow, spectral indexes may not detect them from mid-resolution satellite images.
Hence, a morphological dilation is applied to fill such gaps along the trail of rivers over the outcome of \emph{MNDWI}. 
In dilation, a structural element is used to probe and expand the shapes in an image such that
areas, which are separated by space smaller than the structural element, get connected.  
\emph{MNDWI} detects different classes of water bodies including rivers and lakes.
Water bodies, such as lakes, and swamps, rarely have high mineral content regions.
However, the surrounding of mine swamps has different minerals.
Therefore, rivers are needed to be separated.
To separate rivers from other water bodies, the shapes of these water bodies are studied next.
Image thinning~\cite{zhang1984fast}, which is exerted to obtain the skeleton structures of different shapes by removing black foreground pixels layer by layer, is applied to the dilated outcome.
The thinned versions of rivers are found to be different than other water bodies as rivers are stretched over a larger region, unlike other water bodies.
To separate them, a connected component analysis is applied to the thinned image.
The larger connected components are detected as probable river regions. 
Let these regions be denoted as $R_w$.
This segment is shown in Fig.~\ref{fig:flow} using the grey boxes. 
\subsection{Detection of River Sandbank Regions}
The coal mine index (\emph{CMI}), enhances the concept of clay mineral ratio, which detects regions with high mineral-assemblages.
Lower values of \emph{CMI} detect different land classes of surface coal mine regions~\cite{jit_jstar_coal_mine}.
Other high mineral content regions, such as river sandbanks, can also be detected using a threshold over \emph{CMI}~\cite{jit_mtap_overburden}. 
$CMI<0$ has a higher probability to detect surface coal mine regions~\cite{jit_jstar_coal_mine}.
Hence, a thresholding operation is used 
to detect probable river sandbank regions 
as shown in Eqn~\ref{index_threshold}.
\begin{equation}
\label{index_threshold}
 f(\phi) = \left\{
  \begin{array}{@{}ll@{}}
    ProbableRiverSandbank, & \text{if}\ \emph{CMI} < \tau_1 \: \&\& \: \emph{CMI} > \tau_2\\
    NonRiverSandBank, & \text{otherwise}
  \end{array}\right.
\end{equation}
Here, $\tau_1$, the threshold value, is chosen empirically within $[0.08-0.11]$ and $\tau_2$ is chosen as $0$.
by observing the distribution of \emph{CMI} values in River sandbank regions over the seasons.
Detected regions have river sandbank regions along with a few other regions with high mineral content, which have near similar \emph{CMI} values. 
Let these regions be denoted as $R_m$.
The regions of $R_m$ are analyzed individually using connected component analysis as  
shown in Fig.~\ref{fig:flow} (blue boxes).
Over each connected component, a bounding box is computed with five pixels padding on each side to study the surrounding areas.
The surrounding area of a river sandbank region is a river, unlike other detected regions.
Therefore, each bounding box is checked whether its surrounding regions have any region, which belongs to $R_w$. 
These regions are considered as the detected river sandbank regions.
The proposed technique can detect river sandbank regions using only three bands of \emph{Green}, \emph{SWIR-I}, and \emph{SWIR-II}. It is applicable to any satellite commodity, which has these three bands.
\section{Data and Study Area}
\label{sec:data}
\begin{figure}[ht!]
	\centering
	\includegraphics[width=\columnwidth]{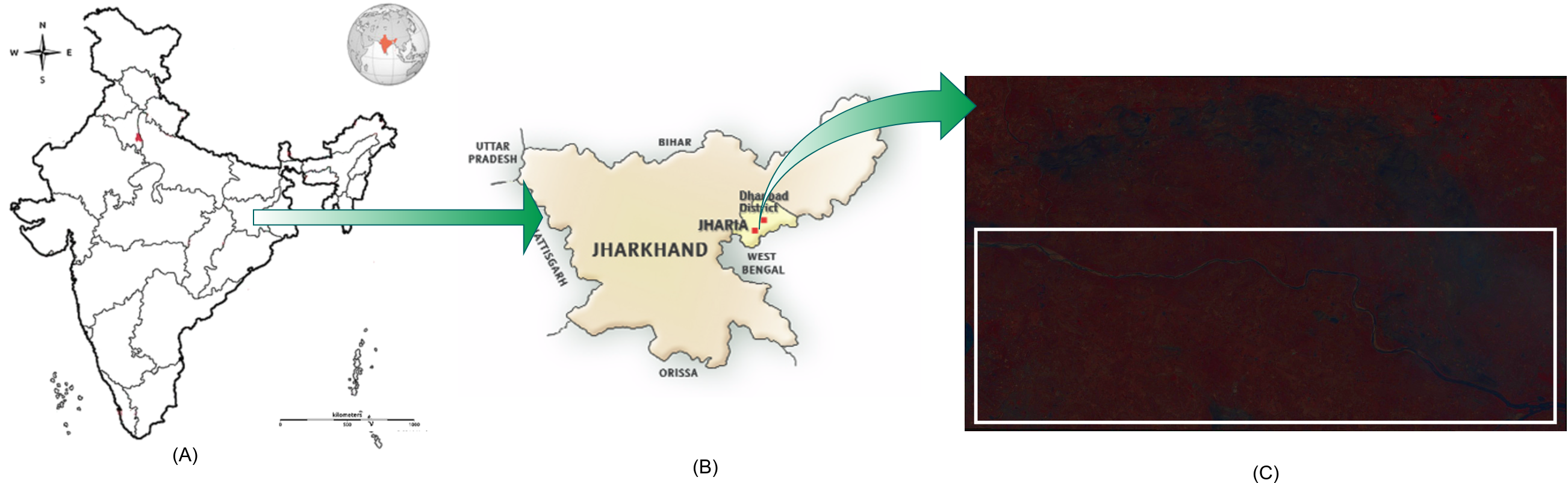}
	\caption{Location of River Sandbank Regions on Damodar River with \emph{JCF}}
	\label{fig:damodar_jharia}
\end{figure}
In this work, the river sandbanks of the Damodar river beside the Jharia Coal Field (\emph{JCF}) region are chosen as the study area as shown in Fig.~\ref{fig:damodar_jharia}. 
The \emph{JCF} is in the Dhanbad district of Jharkhand state in India between latitudes 
$23^{\circ}38'$ N and $23^{\circ}50'$ N and longitudes $86^{\circ}07'E$ and $86^{\circ}30'E$. 
This region has a humid subtropical climate with monsoon rain and
a vast variety of land classes such as reserve forests, dry croplands, freshwater bodies, riverbanks, rivers, grassland, and urban areas.
Land classes associated with coal mining regions, particularly coal overburden dumps have similar spectral properties with river sandbanks~\cite{jit_mtap_overburden}. 
They both have an abundance of minerals 
and are difficult to separate using geophysical indexes.
Hence, this region has been chosen as the study area as it has prominent river sandbanks and coal mining regions both.
Landsat $8$ $L1$ images (Path $140$, Rows $43$ and $44$ as per Landsat Reference System), which produce top of atmosphere (\emph{TOA}) reflectance, in $2017$ are used here.
Landsat $8$ provides nine multi-spectral bands and two thermal bands.
These bands except the panchromatic band have $30$ meter spatial resolution.
These images are radiometrically corrected, and orthorectified.
As multi-spectral images are inapplicable with the presence of clouds, images with less than $10\%$ cloud cover are chosen.
Most of the rainfall in~\emph{JCF} is concentrated from June to the end of September. 
Thus, the proposed work has considered the months from November to May for experimentation.
High-resolution Google Earth images are used here for validation.
Various land classes using these images are marked as ground truth by visual observation.
They are further registered with Landsat $8$ images using geo-reference for accuracy computation. 
A contrast-enhanced false-color image composed of near infra-red (\emph{NIR}), red, and green bands of the study area is shown in Fig.~\ref{fig:damodar_jharia} (C).
Prominent river sandbanks can be visible on the Damodar river at the white bounding box of Fig.~\ref{fig:damodar_jharia} (C).
The scattered darker region above the white bounding box which forms a sickle-like shape is the~\emph{JCF}.
\section{Results and Discussion}
\begin{figure}[ht!]
	\centering
	\includegraphics[width=\columnwidth]{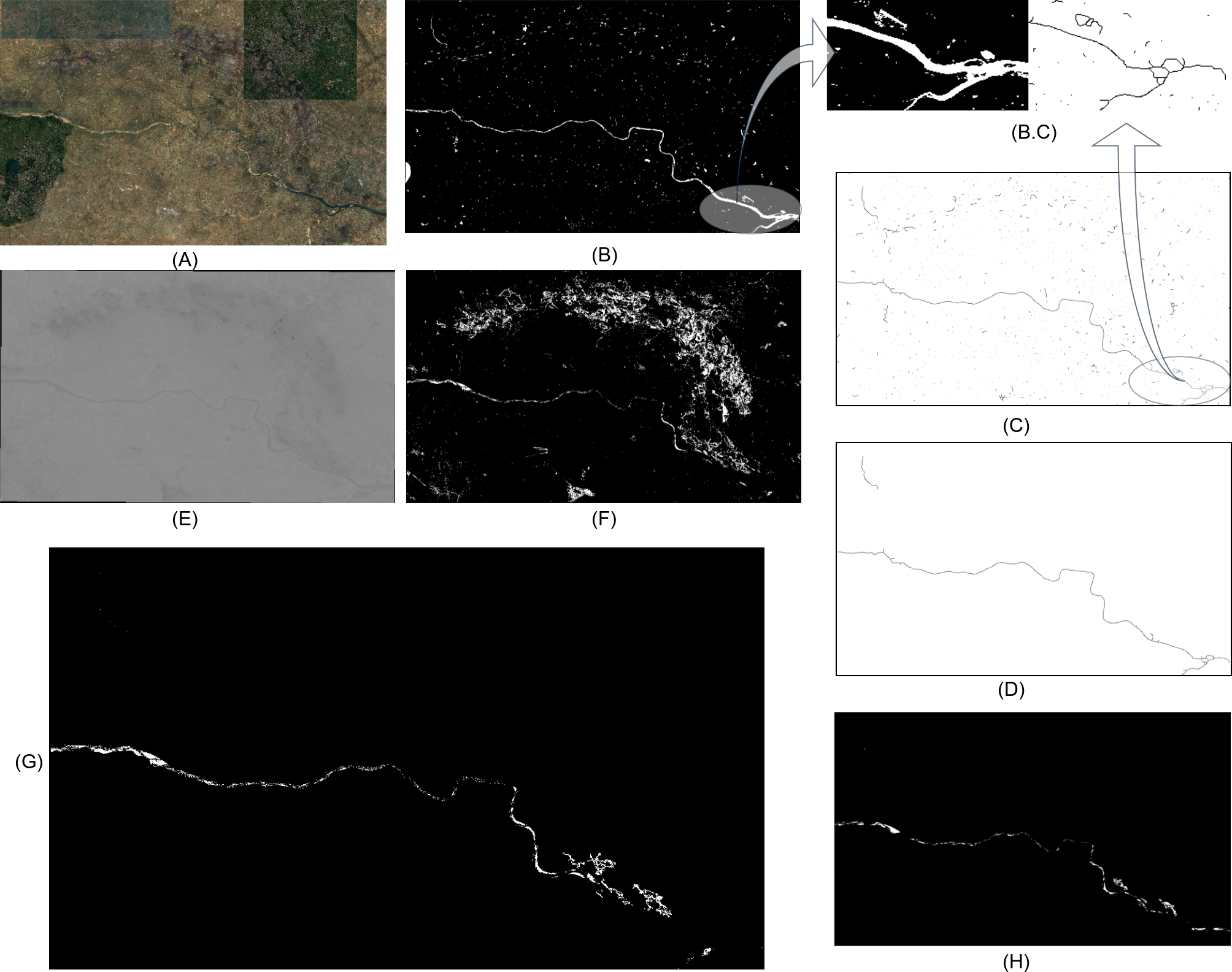}
	\caption{Results: (A) A Ground truth Google Earth Image, (B) Water Bodies, (C) Dilate and Thinning, (B.C) Zoomed in (B) \& (C), (D) Larger Regions by thresholding over (C), (E) \emph{CMI} Image, (F) River Sandbank with Other High Mineral Regions, Detected River Sandbank Regions (G) Landsat (H) Sentinel}
	\label{fig:all_result}
\end{figure}
Landsat 8 images of different months in $2017$ are used here for generating the results and accuracy computation. 
As multispectral images can not trespass clouds, months with heavy rains and $>10\%$ cloud covers in the land are not considered here. 
A consolidated sample outcome of different stages of the proposed algorithm is shown in Fig.~\ref{fig:all_result}.
Fig.~\ref{fig:all_result} (A) shows a high-resolution Google Earth image of the region of interest. 
The proposed technique is an amalgamation of two separate processes.
First, probable river regions are isolated using the morphological patterns of the detected water bodies. 
Thereafter, regions with high mineral contents are detected through the \emph{CMI} values. 
As \emph{MNDWI} has been found beneficial for open water features, which is correlated with the study area, \emph{MNDWI} has been preferred here for water body detection~\cite{xu2006mndwi}.
Fig.~\ref{fig:all_result} (B) shows the outcome of \emph{MNDWI} over the region of interest.
Due to the prominent river sandbank, the river trails may become so narrow that it is incomprehensible in mid-resolution satellite images.
These isolated regions are scattered along the river trails. 
It becomes erroneous if high mineral content regions surrounding these detected river paths are considered to detect river sandbanks.
Hence, morphological dilation is applied over the outcome of \emph{MNDWI}.
Mine swamps and rivers both have high mineral content in the surroundings. 
Thus, image thinning is performed over the dilated outcome to study the shapes of different water bodies (Fig.~\ref{fig:all_result} (C)).
Zoomed versions of \emph{MNDWI} and thinned output are also shown in Fig.~\ref{fig:all_result}.
All the water bodies are treated separately using connected component analysis.
The shape of a thinned river path is found to be larger than the shapes of smaller water bodies, lakes, and mine swamps. 
Hence, larger thinned areas are considered as probable river regions (Fig.~\ref{fig:all_result} (D)). 
Thinned structures of large lakes, dams, and reservoirs can be large enough to get falsely detected as a probable river by this approach. 
However, they rarely have an abundance of mineral-enriched sand in their surroundings for sand mining. 
Therefore, it is observed that such misclassified regions are isolated in the latter part of the proposed technique. 
\emph{CMI} values of the region of interest are shown in Fig.~\ref{fig:all_result} (E).
As discussed in~\cite{jitm2019igarss}, river sandbank regions are observed along with different land classes, such as coal overburden dump regions at a certain stage of the hierarchical clustering tree.  
Additionally,~\cite{jit_mtap_overburden} has noticed a similar spectral pattern of river sandbank and coal overburden dump regions while detecting coal overburden dump. 
~\cite{jit_jstar_coal_mine} detects surface coal mine regions over the season where \emph{CMI} values are $<[0-0.02]$.  
It has been observed that the range of \emph{CMI} value in river sandbank regions is similar to a few land classes such as coal overburden dump regions as shown in Fig.~\ref{fig:all_result} (F). 
However, these regions may or may not have water bodies in surrounding regions. 
It can be observed from Fig.~\ref{fig:all_result} (C) and (F) that there are few water body regions whose surroundings have high mineral content and similar \emph{CMI} values.
The distinguishing factor between these regions and prominent river sandbanks is a river trail.
All the detected regions of Fig.~\ref{fig:all_result} (F) are treated individually using connected component analysis.
A bounding box padded with five pixels on each side over each connected component is considered to check whether it contains a probable river region.
The final outcome of the proposed technique is shown in Fig.~\ref{fig:all_result} (G).  
As observed from Fig.~\ref{fig:all_result}, the proposed technique can detect river sandbank regions with high mineral abundance.
Fig.~\ref{fig:all_result} (H) shows the outcome with Sentinel $2A$ images which have $20m$ resolution.
It is observed that the river sandbank regions can also be detected from Sentinel $2A$ images.
There are a few regions in the proximity of a river, which have high mineral quantity and are not river sandbanks.
The proposed technique can falsely detect them due to the use of a bounding box with five pixels padding (Fig~\ref{fig:all_result} (G), (H)). 
It can be addressed with smaller padding in the future.
Soil has several lithological sub-classes.
In typical cases, the portions of minerals available in sands are higher than different lithological classes present near the rivers such as clay, silt, etc.  
\emph{CMI} preserves the high mineral quantity regions.
A few works study the spectral reflectance properties of different soil classes.
As an example, spectral reflectances of different soil and soil mixtures in different spectrum are computed to check soil fertility in~\cite{patel2022fractional}. 
\emph{SWIR-I} and \emph{SWIR-II} have wavelengths of $1.57-1.67\mu m$ and $2.11-2.29\mu m$, respectively.
It can be observed from~\cite{patel2022fractional} that the difference in reflectance values between \emph{SWIR-I} and \emph{SWIR-II} bands for different soil and soil mixture, are higher than sandy soils.
Furthermore, it has been observed that near infra-red regions ($0.7 - 2.5\mu m$) can differentiate between physical and chemical properties of soils~\cite{rossel2006visible}.
Hence, \emph{CMI} is likely to separate high mineral content sand regions from other lithological classes of soil found near rivers.
More experimentation in such directions is treated as a future work.
\par
\begin{figure}[ht!]
	\centering
	\includegraphics[width=\columnwidth]{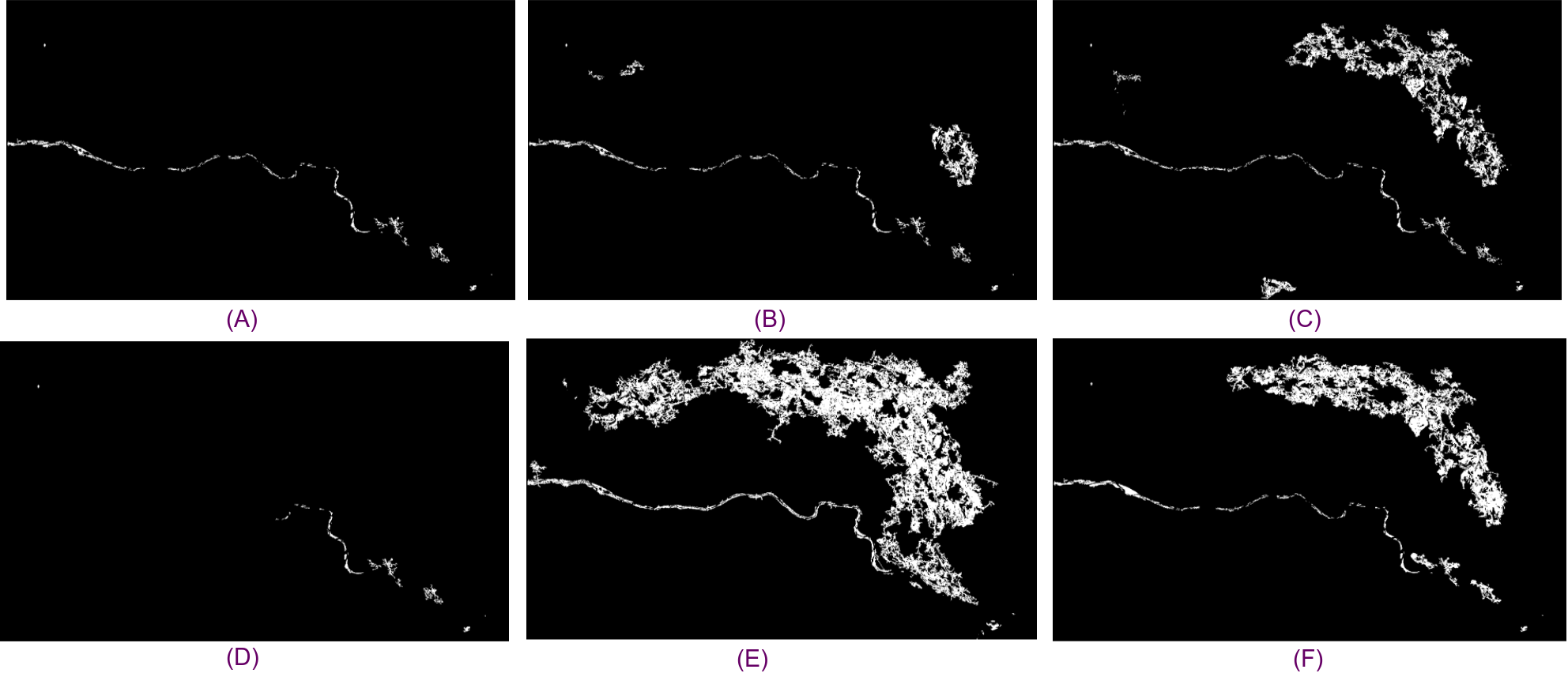}
	\caption{Results with Varying Threshold: (A) Final Output, (B) Lower threshold values for River, (C) Lower Threshold for \emph{MNDWI}, (D) Higher Threshold for \emph{MNDWI}, (E) $\tau_1 > 0.11$, (F) $\tau_2 < 0$}
	\label{fig:thrsh_result}
\end{figure}

The proposed algorithm uses different thresholds values in different phases of its process flow. 
Fig.~\ref{fig:thrsh_result} shows the outcome of the proposed algorithm with varying threshold values at different phases.
Fig.~\ref{fig:thrsh_result} (A) shows the outcome of the proposed algorithm in December 2017. 
The probable river regions are detected after morphological operations and connected component analysis.
A typical river has a longer stretch than other water bodies. 
Thus, connected components with higher regions are selected as the probable river regions because of their longer lengths.
However, if connected components with smaller regions are also preserved, water bodies with smaller length can be falsely get detected as probable rivers. 
As such connected components preserve water bodies with smaller regions, different regions of $R_m$ are falsely detected as river sandbank regions.
Hence, the number of false positive river sandbank regions is increased as shown in Fig.~\ref{fig:thrsh_result} (B).
Higher values of \emph{MNDWI} preserve water bodies.
Lower threshold values of \emph{MNDWI} detects different regions, which are not water bodies.
These falsely detected regions close to a water body, may provide a large connected component.
This directly affects the proposed algorithm as shown in Fig.~\ref{fig:thrsh_result} (C).
Furthermore, if higher threshold values of \emph{MNDWI} is considered, it could not detect different portions of a river with river sandbank. 
Hence, a river appears as isolated segments of trails rather than a continuous one.
The connected component discards many such isolated segments.
In such scenarios, different true river sandbank regions could not get detected as shown in Fig.~\ref{fig:thrsh_result} (D).  
Connected component analysis is also applied over the \emph{CMI} detected regions. 
These regions are detected using two threshold values $\tau_1$ and $\tau_2$. 
Here, $\tau_1$ is chosen empirically within $[0.08-0.11]$ and $\tau_2$ is chosen as $0$.
As it has been observed that $CMI<0$ has higher probability of detecting mining regions~\cite{mukherjee2022study}, $\tau_2$ is chosen as $0$.
The value of $\tau_1$ is chosen empirically by observing the distribution of river sandbank regions over the season.  
If $\tau_1>0.11$ is selected, other regions are also preserved. 
These regions may get connected with each other and form a larger connected component region.
If $\tau_2<0$ is selected various other regions with high mineral assemblages are detected and form a larger connected component.
Such larger connected components have bigger bounding boxes, which cover substantial regions. 
These bounding boxes may have a probable river as detected by the proposed algorithm, where the river is not in the close proximity of the high mineral content region. 
Hence, as shown in Fig.~\ref{fig:thrsh_result} (E) and (F), different false positive regions are also detected.
\subsection{Validation}
\begin{figure}[ht!]
	\centering
	\includegraphics[width=\columnwidth]{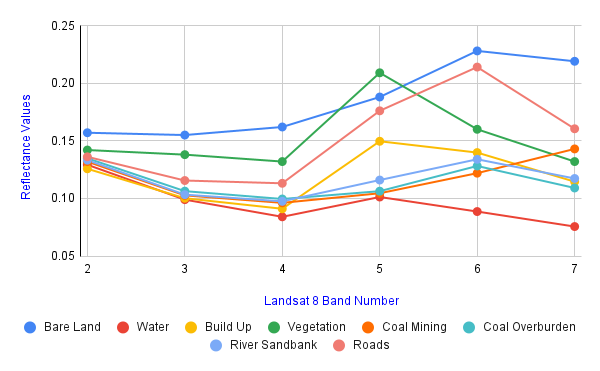}
	\caption{Spectral Responses - Bare land, Water Body, Build Up, Vegetation, River Sandbank, Roads and Coal Mine Land Classes}
	\label{fig:spectral_response}
\end{figure} 
Different land classes are marked using visual interpretation from high-resolution Google Earth images.
These images are used for validation and accuracy computation. 
Primary land classes of the study area, which are considered, are vegetation, water body, bare land, land classes related to surface coal mining, urban regions, and river sandbank. 
These regions are marked and extracted from Landsat 8 and Google earth images using open source \emph{QGIS} software.
Water bodies, which have nearby high mineral content regions are considered.
Mean reflectance values of these ground truth regions over the different spectrum of Landsat 8 are shown in Fig.~\ref{fig:spectral_response}.
$100$ random samples of these ground truth images are analyzed using a t-test with the null hypothesis of $\mu_{RiverSandbank} = \mu_{OtherLandClass}$.
Here, $\mu_{OtherLandClass}$, and $\mu_{RiverSandbank}$ are defined as the mean of \emph{CMI} values of different other ground truth land classes and river sandbank regions, respectively.
Vegetation, water bodies, urban regions, coal mining regions, and coal overburden dump regions are considered as other land classes.
$S_{RiverSandbank} \neq S_{OtherLandClass}$ is considered, where $S$ is defined as the variance.
The alternative hypothesis is defined as $\mu_{RiverSandbank} \neq \mu_{OtherLandClass}$.
The t statistics, degree of freedom, and corresponding p values over the seasons are shown in Table~\ref{tab:tTest}.
As observed in Table~\ref{tab:tTest}, the null hypothesis can be rejected for all the land classes over the season except coal overburden regions. 
Hence, \emph{CMI} can differentiate river sandbank regions from other regions except coal overburden dump over the season.
Similar observations can be found in Fig.~\ref{fig:spectral_response}.
The horizontal, and vertical axes of Fig.~\ref{fig:spectral_response} represent band numbers of Landsat 8, and reflectance values, respectively.
It can be observed that coal overburden dumps have similar spectral characteristics with river sandbank regions (Fig.~\ref{fig:spectral_response}). 
They show near similar values in the Green band (band $3$).
However, coal overburden dumps provide slightly lower values in \emph{NIR} (band $5$), \emph{SWIR-I} (band $6$), and \emph{SWIR-II} (band $7$) bands than river sandbank (Fig.~\ref{fig:spectral_response}). 
The difference in the \emph{NIR} band is higher than the other two bands (Fig.~\ref{fig:spectral_response}).
It corroborate with the findings in~\cite{jit_mtap_overburden}.
Therefore, as observed in Fig.~\ref{fig:spectral_response}, though water indexes can separate coal overburden dump from river sandbank regions, but other isolated land classes get detected along with river sandbank regions.
Hence, in this paper, the presence of rivers in the nearby regions is preferred rather than using water indexes exclusively.
\begin{table}[]
\caption{T-Test Result on \emph{CMI} Values over the Null Hypothesis $\mu_{RiverSandbank} = \mu_{OtherLandClass}$}
\centering
\begin{tabular}{|ll|l|l|l|l|l|}
\hline
\multicolumn{2}{|l|}{}                                 & Vegetation  & Water       & Urban       & \begin{tabular}[c]{@{}l@{}}Coal\\ Mining\end{tabular} & \begin{tabular}[c]{@{}l@{}}Coal \\ Overburden\end{tabular} \\ \hline
\multicolumn{1}{|l|}{\multirow{3}{*}{Nov}}   & $t_0$   & $-37.3$     & $-5.59$     & $-13.96$    & $17.97$                                               & $-4.37$                                                    \\ \cline{2-7} 
\multicolumn{1}{|l|}{}                       & $df$    & $137.6$     & $104.6$     & $156.7$     & $99.2$                                                & $118..7$                                                   \\ \cline{2-7} 
\multicolumn{1}{|l|}{}                       & P Value & $< 0.00001$ & $< 0.00001$ & $< 0.00001$ & $< 0.00001$                                           & $0.00003$                                                  \\ \hline
\multicolumn{1}{|l|}{\multirow{3}{*}{Dec}}   & $t_0$   & $-23.85$    & $-3.21$     & $-12.77$    & $15.97$                                               & $-1.09$                                                    \\ \cline{2-7} 
\multicolumn{1}{|l|}{}                       & $df$    & $144.5$     & $116.6$     & $170.1$     & $99.1$                                                & $125$                                                      \\ \cline{2-7} 
\multicolumn{1}{|l|}{}                       & P Value & $< 0.00001$ & $0.001715$  & $< 0.00001$ & $< 0.00001$                                           & $0.27781$                                                  \\ \hline
\multicolumn{1}{|l|}{\multirow{3}{*}{Jan}}   & $t_0$   & $-17.08$    & $-2.62$     & $-7.11$     & $13.79$                                               & $5.05$                                                     \\ \cline{2-7} 
\multicolumn{1}{|l|}{}                       & $df$    & $166$       & $107.8$     & $195.3$     & $99$                                                  & $141.1$                                                    \\ \cline{2-7} 
\multicolumn{1}{|l|}{}                       & P Value & $< 0.00001$ & $0.010061$  & $< 0.00001$ & $< 0.00001$                                           & $< 0.00001$                                                \\ \hline
\multicolumn{1}{|l|}{\multirow{3}{*}{Feb}}   & $t_0$   & $-11.3$     & $-3.52$     & $-4.5$      & $12.62$                                               & $3.83$                                                     \\ \cline{2-7} 
\multicolumn{1}{|l|}{}                       & $df$    & $199.7$     & $121.3$     & $180$       & $99$                                                  & $193.6$                                                    \\ \cline{2-7} 
\multicolumn{1}{|l|}{}                       & P Value & $< 0.00001$ & $0.000609$  & $0.000012$  & $< 0.00001$                                           & $0.000173$                                                 \\ \hline
\multicolumn{1}{|l|}{\multirow{3}{*}{March}} & $t_0$   & $-10.5$     & $-3.27$     & $-5.62$     & $22.45$                                               & $6.2$                                                      \\ \cline{2-7} 
\multicolumn{1}{|l|}{}                       & $df$    & $191.6$     & $129.3$     & $199.9$     & $99$                                                  & $174$                                                      \\ \cline{2-7} 
\multicolumn{1}{|l|}{}                       & P Value & $< 0.00001$ & $0.001378$  & $< 0.00001$ & $< 0.00001$                                           & $< 0.00001$                                                \\ \hline
\multicolumn{1}{|l|}{\multirow{3}{*}{May}}   & $t_0$   & $-15.78$    & $-9.65$     & $-7.83$     & $51.96$                                               & $2.39$                                                     \\ \cline{2-7} 
\multicolumn{1}{|l|}{}                       & $df$    & $154$       & $148.4$     & $165.4$     & $99$                                                  & $131.5$                                                    \\ \cline{2-7} 
\multicolumn{1}{|l|}{}                       & P Value & $< 0.00001$ & $< 0.00001$ & $< 0.00001$ & $< 0.00001$                                           & $0.018267$                                                 \\ \hline
\end{tabular}
\label{tab:tTest}
\end{table}
\par
\begin{figure}[ht!]
	\centering
	\includegraphics[width=\columnwidth]{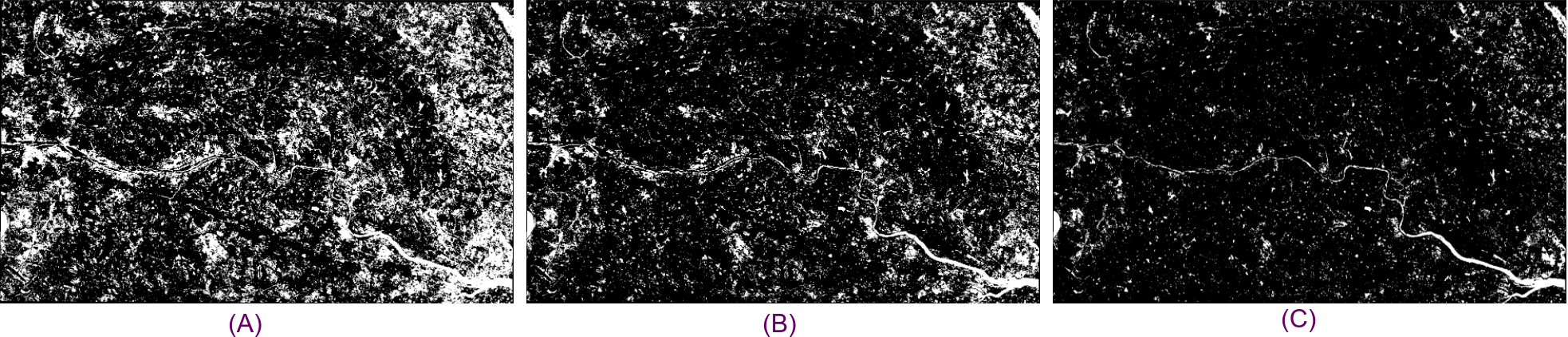}
	\caption{Results: \emph{NDMI} with threshold values of (A) $0.1$, (B) $0.15$, (C) $0.2$}
	\label{fig:NDMI}
\end{figure}

Most of the roads found in this region are insignificant as per the spatial resolution of Landsat $8$. 
The proposed technique may be found erroneous when there are prominent roads and \emph{MNDWI} fails to separate them from water bodies. 
Thus, the spectral responses of roads are computed (Fig.~\ref{fig:spectral_response}). 
It has been observed that water bodies have significantly lower \emph{SWIR-I} values than \emph{NIR}, whereas roads may not follow this trend. 
This characteristic is also observed in~\cite{kumar2021index} while detecting roads. 
Normalized difference moisture index (\emph{NDMI}) is defined as $\frac{\lambda_{NIR}-\lambda_{SWIR-I}}{\lambda_{NIR}+\lambda_{SWIR-I}}$. 
Next, the mean \emph{NDMI} value ($\mu_{NDMI}$) of each connected component is checked.
If $\mu_{NDMI} \ngtr \xi$, that connected component is discarded.
Here, $\xi$ is chosen empirically as $0.15$ as shown in Fig.~\ref{fig:NDMI}.
Fig.~\ref{fig:NDMI} shows outcome of \emph{NDMI} with different threshold values at the region of interest.
As shown in Fig.~\ref{fig:NDMI} (C), all the water bodies are primarily detected when the threshold value is $0.2$.
However, a few portions of the river network remain undetected.
When the threshold values is chosen as $0.1$, various other regions including a few of the road networks are also detected as shown in Fig.~\ref{fig:NDMI} (A).
Though different other small regions are detected along with the water bodies, most of the river network is detected when threshold value is chosen as $0.15$ (Fig.~\ref{fig:NDMI} (B)).
Thus, here, $\xi$ is chosen empirically as $0.15$.
Thorough experimentation regarding this is treated as a future work.
The separation of water bodies and road network can be improved by any robust technique to detect roads from satellite images.
Different supervised machine learning based techniques can be a probable solution.
\subsection{Performance Analysis}
\begin{figure}[ht!]
	\centering
	\includegraphics[width=\columnwidth]{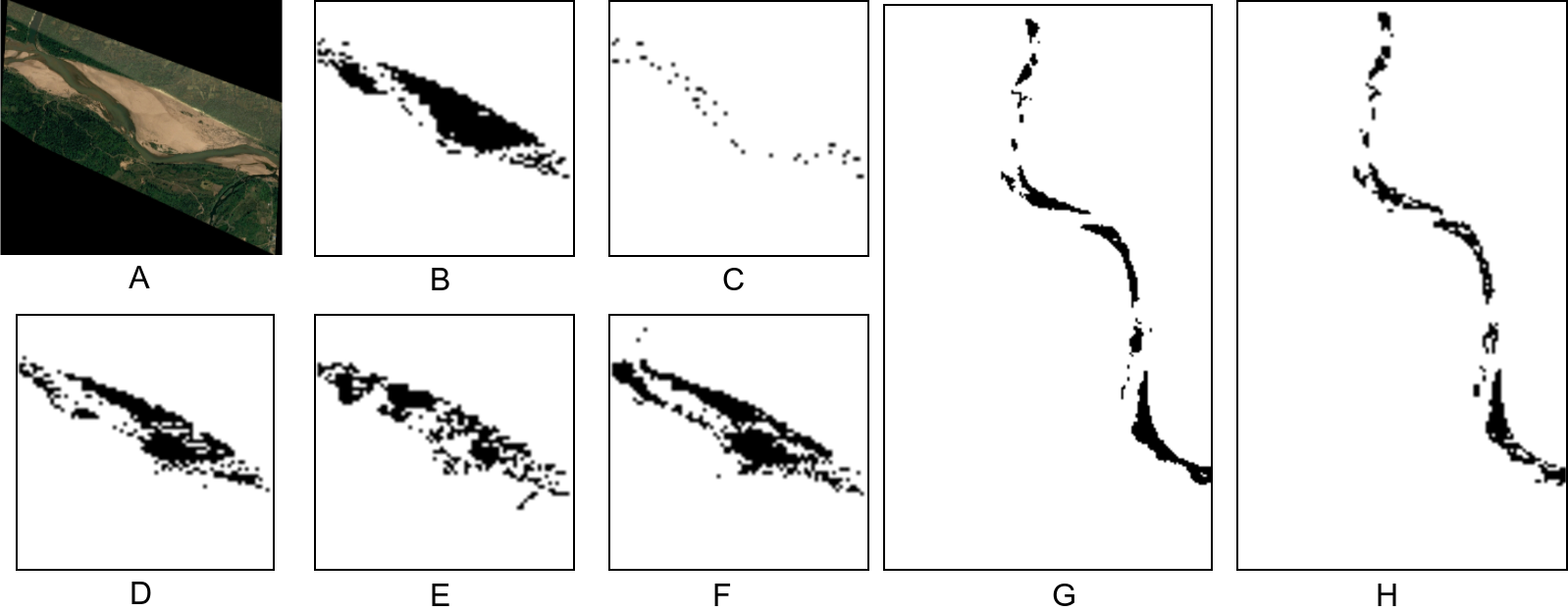}
	\caption{(A) Ground truth, Regions: (B) Detected, (C) Falsely Detected (2017); Effect of Rain (D) April, (E) Oct, (F) Dec, 2021; (G) \emph{Sentinel} (H) \emph{Landsat}}
	\label{fig:result_accuracy}
\end{figure}
\begin{table}[]
\caption{Accuracy Computation of the Proposed Technique}
\begin{center}
\begin{tabular}{|c|c|c|c|c|c|c|}
\hline
           & Dec       & Jan       & Feb       & March     & May       & Nov       \\ \hline
Precision (\%)   & $91.85$ & $85.32$ & $87.2$  & $73.6$  & $84.2$  & $90.66$ \\ \hline
Recall  (\%)   & $79.23$ & $70.6$  & $69.64$ & $75.71$ & $69.8$  & $76.03$ \\ \hline
$F_1$ Score (\%) & $85.07$ & $77.27$ & $77.44$ & $74.64$ & $76.33$ & $82.71$ \\ \hline
Accuracy  (\%)  & $93.17$ & $90.3$  & $90.51$ & $88.25$ & $89.94$   & $92.35$ \\ \hline
\end{tabular}
\end{center}
\label{tab_accuracy}
\end{table}
These ground truth regions are utilised for the accuracy computation.
The proposed technique has average precision, recall, $F_1$ score, and accuracy of $85.47\%$, $73.5\%$, $78.91\%$, and $90.75\%$, respectively over the seasons as shown in Table~\ref{tab_accuracy}.
Fig.~\ref{fig:result_accuracy} shows an outcome of the proposed technique in a smaller region.
Fig.~\ref{fig:result_accuracy} (A), and (B) show the google earth image and the outcome of the proposed method, respectively.
Fig.~\ref{fig:result_accuracy} (C) shows the regions which are falsely detected as river sandbank regions. 
It can be observed from Fig.~\ref{fig:result_accuracy} that the proposed technique can detect the river sandbank regions.  
The increased water content in rainy seasons directly affects the river sandbank regions as shown in Fig.~\ref{fig:result_accuracy} (D-F). 
Fig.~\ref{fig:result_accuracy} (D), (E), and (F) show the outcomes of the proposed technique before, just after, and after rainy seasons, respectively, in $2021$. 
Due to the high water content of the river, it can be observed from Fig.~\ref{fig:result_accuracy} ($D-F$) that the shape of the river sandbank regions has been changed and the proposed technique can detect them.
As observed in Fig.~\ref{fig:result_accuracy}, and Table~\ref{tab_accuracy}, the proposed technique can identify the river sandbank regions with good precision.
The segments of outcomes in Sentinel $2A$, and Landsat $8$ data are shown in Fig.~\ref{fig:result_accuracy}(G), and (H), respectively.
The proposed technique shows higher precision of $88.79\%$, and higher recall of $78.68\%$ by using Sentinel $2A$.
As the boundary of the detected regions is smoother in Sentinel images (Fig.~\ref{fig:result_accuracy} (G)), the accuracy of the proposed technique may improve if finer spatial resolution is used.
It can be observed from Table~\ref{tab_accuracy} that the proposed technique has comparatively low accuracy and $F_1$ score in March and May than other months.
\emph{CMI} has been used here to detect high mineral content regions. 
In has been observed that \emph{CMI} shows comparatively lower accuracy of in summer seasons than other seasons~\cite{jit_jstar_coal_mine,jit_mtap_overburden}.
In this work, \emph{CMI} and \emph{MNDWI} have been used. 
Both of them use short wave infra-red bands, which have also been used to compute moisture content in the past.
As summer months are dry months, this may have affected the performance due to the variance in soil moisture~\cite{jit_jstar_coal_mine}.
Thus, it may have an adverse affect on the accuracy of the proposed algorithm in summer.
Further experimentation in these regards is treated as a future work.
\par
\begin{figure}[ht!]
	\centering
	\includegraphics[width=\columnwidth]{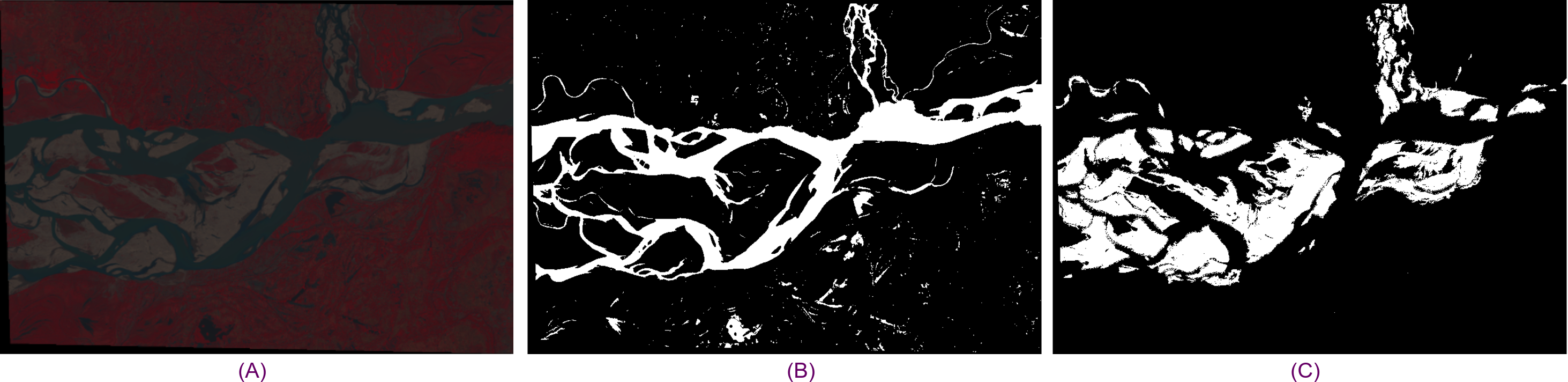}
	\caption{Results with Bramhaputra River in Assam, India: (A) False Color Image, (B) Detected Water Bodies, (C) Detected River Sandbank}
	\label{fig:otherPlace_asam}
\end{figure}
\begin{figure}[ht!]
	\centering
	\includegraphics[width=\columnwidth]{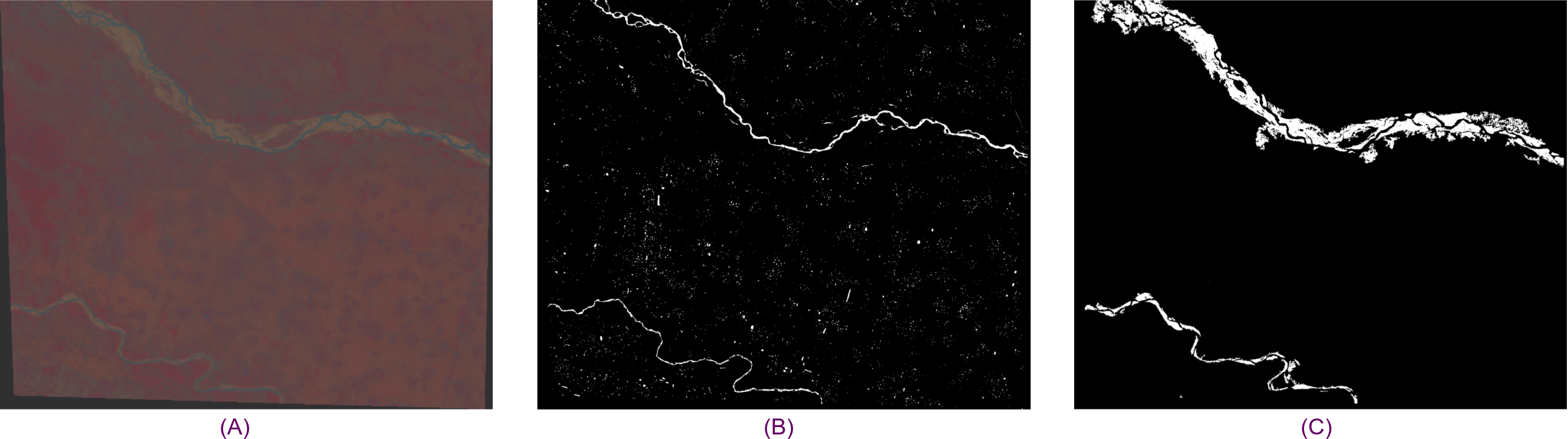}
	\caption{Results with Damodar and Dwarakeswar Rivers in West Bengal, India: (A) False Color Image, (B) Detected Water Bodies, (C) Detected River Sandbank}
	\label{fig:otherPlace_damodar}
\end{figure}

The proposed technique has been applied in different other regions.
Fig.~\ref{fig:otherPlace_asam} shows the outcome of the proposed technique over the Bramhaputra river close to Guwahati, Assam, India.
A false color representation of the region of interest, detected water bodies, and detected river sandbank regions are shown in Fig.~\ref{fig:otherPlace_asam} (A), (B), and (C), respectively.
The Bramhaputra river shows significant sediment discharge and has average width of $5.46km$~\cite{assam_bramhaputra}. 
The regions shows tropical monsoon rainforest climate. 
It can be observed from Fig.~\ref{fig:otherPlace_asam} that the proposed technique can detect multiple, complex and variable length river sandbank regions in a large river system.
The proposed technique has been also tested over the Damodar and Dwarakeswar river near to to Burdwan, West Bengal, India as shown in Fig.~\ref{fig:otherPlace_damodar}. 
Damodar and Dwarakeswar basins are widely used for sand mining.  
A false color representation of the region of interest, detected water bodies, and detected river sandbank regions are shown in Fig.~\ref{fig:otherPlace_damodar} (A), (B), and (C), respectively.
It can be seen from Fig.~\ref{fig:otherPlace_damodar} that the proposed technique can detect river sand bank regions on both the rivers. 
Therefore, the proposed technique can detect river sandbank regions on different rivers together. 
However, a few regions are falsely detected by the proposed technique  in the Dwarakeswar basin. 
As Dwarakeswar river has a narrow width and shows high sinuosity, a few nearby regions may get falsely detected due the bounding box computed over each connected components.
More investigation with such a narrow river is considered as a future work.
\par 
A multi-modal analysis composed of multi-spectral and \emph{InSAR} images is employed to detect ex-sand mining area~\cite{indriasari2018identification}.
Integration of differential global positioning systems and \emph{UAV} data is used to quantify sand quarry volumes~\cite{Stalin2020}.
As \emph{InSAR} and \emph{DEM} data require multiple stages of pre-processing and are not acquired in a regular interval, in most cases,
Landsat $8$ has been used here to detect river sandbank regions which are highly dynamic and need to be observed in regular intervals.
Additionally, different multimodal techniques require the availability of such satellite commodities, which is difficult for several regions.
In semi-supervised or supervised techniques, labeled dataset is required, which may be found challenging for such smaller land classes. 
As an example, deep learning based techniques are employed in~\cite{balaniuk2020mining} to detect mining and tailing dams.
In~\cite{chen2020fine}, different land classes in a mining regions are detected using supervised support vector machine.
Such techniques have the prerequisites of a labeled dataset for training, which may be found difficult for a finer land class.
Moreover, such techniques do not emphasize the distinguishing spectral characteristics of a land class, viz. river sandbanks.
The proposed method provides a novel technique to detect river sandbank 
regions in the presence of other regions containing ample amounts of minerals using multi-spectral images without any labeled dataset.
Although some of the river sandbanks can be found visually separable, the abundance of minerals is difficult to comprehend using \emph{Red-Green-Blue} bands, which the proposed technique addresses. 
Landsat $8$ provides mid-resolution satellite images with a spatial resolution of $30m$. 
It is difficult to detect those river sandbank regions, which are smaller than the spatial resolution of Landsat 8 satellite images.
However, Landsat images have been used in the literature to detect river sandbank regions. 
As an example, in~\cite{indriasari2018identification}, deformation of ex-mining areas of sand is detected using Landsat $5$ images (spatial resolution of $30m$) and \emph{DEM} data derived from phased array type L-band synthetic aperture radar (\emph{PALSAR}) of advanced land observing satellite (\emph{ALOS}). 
Green and Red channels of Landsat $8$ images are used to quantify suspended
sediment concentration in Red River~\cite{pham2018using}. 
The proposed technique has also been found applicable to Sentinel $2A$ images.
Detection of sandy regions, which have ample amounts of minerals and  are suitable for sand mining, is challenging. 
Landsat 8 provides various bands, which are susceptible to high mineral-assemblages. 
Hence, in this work, Landsat $8$ images are considered. 
\par
River sandbank regions are finer land classes and hence, they are difficult to detect.
Generating manually labeled dataset of such a finer land class is laborious and error-prone.
These land classes are dynamic and their spectral trends can be affected by the season.
However, the proposed technique provides a novel technique to detect river sandbank regions over the seasons without labeled dataset. 
Thus, the proposed technique can be have wider applications irrespective of season and available labeled dataset.
A time-series analysis using the proposed technique can be used to quantify the changes of river sandbank and their probable effect on the river morphology.
The proposed technique can be found useful for identification of the illegal sand mining, which has drastic measures on river health.
The inclusion of different characteristics of river trails, such as sinuosity, and \emph{DEM} can boost the performance of river trail detection, which is treated as one of the future directions of this work.
Here, results are generated using \emph{TOA} reflectance from $L1$ data. 
\emph{CMI} and \emph{MNDWI}, both are found to be robust to surface reflectance~\cite{jit_jstar_coal_mine, xu2006mndwi}.
Further experimentation with $L2$ data and other climatic regions is considered as a future reference. 
\section{Conclusion}
River sandbank is one of the primary sources of sand mining and has near similar characteristics to 
subclasses of bare soil regions, which makes it difficult to separate.
The proposed work presents a novel technique to detect river sandbank regions for sand mining using the presence of a high quantity of minerals and association to rivers, without any labeled dataset, and using multi-spectral images exclusively.
Initially, probable river regions are detected.
Next, high mineral content regions are considered.
Regions, which are close to the proximity of rivers and have a high abundance of minerals are detected as river sandbank regions.
It may become misleading with the presence of large water bodies with high mineral contents, which is treated as a future direction of the work. 
However, these regions have less amount of minerals in contrast to rivers. 
Therefore, the final outcome gets less affected by these erroneous regions.
It provides average accuracy, precision, and recall of $90.75\%$, $85.47\%$, and $73.5\%$, respectively over the seasons.
As the proposed technique is dependent on the accuracy of \emph{CMI} and \emph{MNDWI}, it can be further improved by enhancing their performances, which is considered a future work.
The proposed method has significant potentials in different future applications, including, the detection, classification, and monitoring of river sandbank and river morphology, monitoring of riverbed stability, early prediction of river course change, and others.   
It has also several social and industrial applications in the mining industry, resource management, and detection of illegal sand mining.


\bibliography{riversandbank.bib}   
\bibliographystyle{spiejour}   


\vspace{2ex}\noindent\textbf{First Author} is currently an assistant professor at the Computer Science and engineering department of Birla Institute of Technology, Mesra, India. He received his B. Tech degree from West Bengal University of Technology in Computer Science and Engineering. He received his MS degrees in School of Information Technology from the Indian Institute of Technology Kharagpur, India in $2014$. He concluded his doctoral study in remote sensing from the Indian Institute of Technology Kharagpur in $2020$. He has authored more than fifteen journal and conference papers. His current research interests include muti-spectral imaging, image processing, and machine learning.

\vspace{1ex}
\noindent Biographies and photographs of the other authors are not available.

\listoffigures
\listoftables

\end{document}